\documentclass[fleqn]{llncs}
\usepackage{amsmath,amssymb}
\usepackage{pstricks,pst-node}
\usepackage{wrapfig}

\newcommand{\bfemph}[1]         {{\bfseries\boldmath\emph{#1}}}

\newcommand{\mi}[1]             {\mathit{#1}}
\newcommand{\ms}[1]             {\mathsf{#1}}

\newcommand{\lequiv}            {\leftrightarrow}
\newcommand{\limplies}          {\rightarrow}
\newcommand{\qlimplies}         {\quad\rightarrow\quad}

\newcommand{\lAnd}              {\bigwedge}

\newcommand{\setunion}          {\cup}

\newcommand{\setintersection}   {\cap}

\newcommand{\setc}[2]           {\{\, #1 \mid #2 \,\}}
\newcommand{\qtext}[1]          {\quad\text{#1}\quad}

\newcommand{\tuple}[1]          {\langle #1 \rangle}
\newcommand{\rangebopd}[5]      {#1_#2#4 #5 #4#1_#3}
\newcommand{\rangeb}[3]         {\rangebopd{#1}{#2}{#3},{\ldots}}

\newcommand{\rangen}[1]         {\rangebopd{#1}1n,{\ldots}}

\renewcommand{\leq}             {\leqslant}
\renewcommand{\geq}             {\geqslant}

\newcommand{\ie}                {i.e.}
\newcommand{\eg}                {e.g.}

\newcommand{\disjoint}          {\mathsf{disjoint}}
\newcommand{\meet}              {\mathsf{meet}}
\newcommand{\overlap}           {\mathsf{overlap}}
\newcommand{\equal}             {\mathsf{equal}}
\newcommand{\covers}            {\mathsf{covers}}
\newcommand{\coveredby}         {\text{\normalfont\textsf{covered-by}}}
\newcommand{\contains}          {\mathsf{contains}}
\newcommand{\inside}            {\mathsf{inside}}

\newcommand{\north}             {\mathsf{N}}
\newcommand{\northwest}         {\mathsf{NW}}
\newcommand{\northeast}         {\mathsf{NE}}
\newcommand{\south}             {\mathsf{S}}
\newcommand{\southwest}         {\mathsf{SW}}
\newcommand{\southeast}         {\mathsf{SE}}
\newcommand{\east}              {\mathsf{E}}
\newcommand{\west}              {\mathsf{W}}
\newcommand{\samepoint}         {\mathsf{EQ}}

\newcommand{\raa}               {\ \rightarrow\ }

\newcommand{\so}[1]             {\mbox{\textsl{#1}}} 

\newcommand{\Q}                 {\mathcal{Q}}
\newcommand{\OO}                {\mathcal{O}}
\newcommand{\PP}                {\mathcal{P}}
\newcommand{\R}                 {\mathcal{R}}

\newcommand{\Rel}               {\mathit{Rel}}
\newcommand{\CT}                {\mathit{CT}}

\newcommand{\RCC}               {\mathsf{RCC8}}
\newcommand{\Dir}               {\mathsf{Dir}}

\newcommand{\comp}              {\mathsf{comp}}
\newcommand{\conv}              {\mathsf{conv}}
\newcommand{\neighbour}         {\mathsf{neighbour}}
\newcommand{\link}              {\mathsf{link}}

\newcommand{\dom}               {\mathit{dom}}

\newcommand{\eclipse}           {\textup{ECL\textsuperscript{\textit{i}}PS%
                                \textsuperscript{\textit{e}}}}

\newcommand{\cons}[2]           {\mathit{cons}(#2, #1)}
\newcommand{\consb}[3]          {\cons{#1}{#2} \equiv #3}

\newcommand{\setnewlength}[2]   {\newlength{#1}\setlength{#1}{#2}}

\setnewlength{\mnl}{0.666ex}
\setnewlength{\mindent}{2em}

\setnewlength{\algotab}{2.8em}
\setnewlength{\algotabinitial}{1em}
\setnewlength{\algotabmargin}{1em}
\setnewlength{\algogaplen}{1ex}

        {\end{tabbing}}

\newcommand{\always}            {{\psset{unit=0.07em,linewidth=0.5}\begin{pspicture}(11,10)\psframe[linearc=0.01](1,0)(10,10)\end{pspicture}}}
\newcommand{\eventually}        {{\psset{unit=0.07em,linewidth=0.5}\begin{pspicture}(11,10.2)\psdiamond[linearc=0.01](5,5)(5.1,5.1)\end{pspicture}}}
\newcommand{\nexttime}          {{\psset{unit=0.07em,linewidth=0.5}\begin{pspicture}(11,10)\pscircle(5,5){5}\end{pspicture}}}
\newcommand{\until}             {\mathbin{\mathsf{U}}}

\newcommand{\zerowidth}[1]      {\makebox[0mm][l]{#1}}

\begin{document}

\title{Infinite Qualitative Simulations \\ by Means of Constraint Programming}
\author{Krzysztof R. Apt\inst{1,2} \and Sebastian Brand\inst{3}}
\institute{CWI, P.O. Box 94079, 1090 GB Amsterdam, the Netherlands\\
\and
University of Amsterdam, the Netherlands
\and
NICTA, Victoria Research Lab, Melbourne, Australia}

\maketitle
\thispagestyle{empty}

\begin{abstract}
  We introduce a constraint-based framework for studying infinite qualitative
  simulations concerned with contingencies such as time, space, shape,
  size, abstracted into a finite set of qualitative relations.  To
  define the simulations we combine constraints that formalize the
  background knowledge concerned with qualitative reasoning with
  appropriate inter-state constraints that are formulated using linear
  temporal logic.

  We implemented this approach in a constraint programming system
  (\eclipse) by drawing on the ideas from bounded model checking.  The
  implementation became realistic only after several rounds of
  optimizations and experimentation with various heuristics.

  The resulting system allows us to test and modify the problem
  specifications in a straightforward way
  and to combine various knowledge aspects.
  To demonstrate the expressiveness and
  simplicity of this approach we discuss in detail two
  examples: a navigation problem and a simulation of juggling.
\end{abstract}


\section{Introduction}

\subsection{Background}

\bfemph{Qualitative reasoning} was introduced in AI
to abstract from numeric quantities, such as the precise
time of an event, or the location or trajectory of an
object in space, and to reason instead on the level
of appropriate abstractions.
Two different forms of qualitative reasoning were studied.  The first
one is concerned with reasoning about continuous change in physical
systems, monitoring streams of observations and simulating behaviours,
to name a few applications.  The main techniques used are qualitative
differential equations, constraint propagation and discrete state
graphs.  For a thorough introduction see
\cite{kuipers:1994:qualitative}.

The second form of qualitative reasoning focuses on the study of
contingencies such as time, space, shape, size, directions, through an
abstraction of the quantitative information into a finite set of
qualitative relations. One then relies on complete knowledge about the
interrelationship between these qualitative relations.  This approach
is exemplified by temporal reasoning due to
\cite{allen:1983:maintaining}, spatial reasoning introduced in
\cite{egenhofer:1991:reasoning} and \cite{randell:1992:computing},
reasoning about cardinal directions (such as North, Northwest); see,
\eg, \cite{ligozat:1998:reasoning}, etc.

In this paper we study the second form of qualitative
reasoning.  Our aim is to show how infinite qualitative simulations
can be naturally formalized by means of temporal logic and constraint
satisfaction problems.
Our approach allows us to use generic constraint programming systems
rather than specialized qualitative reasoning systems.
By a \bfemph{qualitative simulation} we mean a
reasoning about possible evolutions in time of models capturing
qualitative information.  One assumes that time is discrete and that
only changes adhering to some desired format occur at each stage.
Qualitative simulation in the first framework is discussed in
\cite{kuipers:2001:encyclopedia}, while qualitative spatial simulation
is considered in \cite{cui:1992:qualitative}.

\subsection{Approach}

In the traditional constraint-based approach to qualitative reasoning
the qualitative relations (for example $\overlap{}$)
are represented as constraints over
variables with infinite domains (for example closed subsets of
$\R^2$) and path-consistency is used as the constraint
propagation; see, e.g., \cite{escrig:1998:qualitative}.

In our approach we represent qualitative relations \emph{as
  variables}.  This allows us to trade path-consistency for hyper-arc
consistency which is directly available in most constraint programming
systems, and to combine in a simple way various theories constituting the
\emph{background knowledge}.
In turn, the \emph{domain specific knowledge}
about simulations is formulated using the
linear temporal logic.  These temporal formulas are
subsequently translated into constraints.

Standard techniques of constraint programming combined with
techniques from bounded model checking
can then be used to generate simulations.  To
support this claim, we implemented this approach in the constraint
programming system \eclipse{}.
However, this approach became realistic
only after fine-tuning of the translation of temporal formulas to
constraints and a judicious choice of branching strategy and
constraint propagation.
To show its usefulness we discuss in detail two case studies. In each
of them the solutions were successfully found by our implementation,
though for different problems different heuristic had to be used.

The program is easy to use and to interact with.  In fact, in
some of the case studies we found by analyzing the generated solutions
that the specifications were incomplete.  In each case, thanks to the
fact that the domain specific knowledge is formulated using temporal
logic formulas, we could add the missing specifications in a
straightforward way.

\subsection{Structure of the paper}

In Section~\ref{sec:rel-approach}
we discuss examples of
qualitative reasoning and in Section~\ref{sec:var-approach}
explain our formalization of the qualitative reasoning by means of constraints.
Next, in Section~\ref{sec:constraints} we deal
with qualitative simulations by
introducing inter-state constraints which connect different stages of
simulation and determine which scenarios are allowed.
These constraints are defined using linear temporal logic.  Their
semantics is defined employing the concept of a cyclic path borrowed
from the bounded model checking approach (see \cite{biere:2003:bounded})
for testing validity of
temporal formulas.

In Section~\ref{sec:translations} we explain how the inter-state
constraints are translated to constraints of the underlying background
knowledge.  Next, in Section~\ref{sec:implementation} we discuss
technical issues pertaining to our implementation that
generates infinite qualitative simulations.
In the subsequent two sections we report on our case studies.
Finally, in Section~\ref{sec:conclusions} we discuss the related work.


\section{Qualitative Reasoning: Setup and Examples}
\label{sec:rel-approach}

As already said, in qualitative reasoning, one abstracts from the
numeric quantities and reasons instead on the level of their
abstractions. These abstractions are provided in the form of a finite
\bfemph{set of qualitative relations}, which should be contrasted with
the infinite set of possibilities available at the numeric level.
After determining the `background knowledge' about these qualitative
relations we can derive conclusions on an abstract level that would be
difficult to achieve on the numeric level.
The following three examples illustrate the matters.

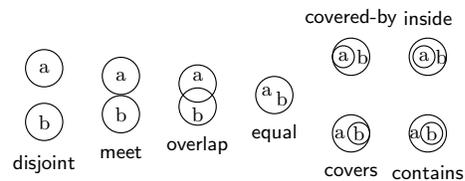
\begin{wrapfigure}[11]{r}[0mm]{62mm}
\vspace*{-3ex}
\centering
\scalebox{0.85}{\small\psset{unit=6mm}
\begin{pspicture}(0,0.5)(12,5.5)
\psset{arrows=<->,nodesep=0.2,linewidth=0.5pt}

\pscircle(1,3.7){0.5}\rput(1,3.7){a}
\pscircle(1,2.3){0.5}\rput(1,2.3){b}
\pnode(1.3,3){disjoint}\rput(1,1.2){$\disjoint$}

\pnode(2.6,3){meet1}
\pscircle(3,3.5){0.5}\rput(3,3.5){a}
\pscircle(3,2.5){0.5}\rput(3,2.5){b}
\pnode(3.4,3){meet2}\rput(3,1.5){$\meet$}

\pnode(4.5,3){overlap1}
\pscircle(5,3.3){0.5}\rput(5,3.5){a}
\pscircle(5,2.7){0.5}\rput(5,2.5){b}
\pnode(5.5,3){overlap2}\rput(5,1.7){$\overlap$}

\cnode(7,3){0.5}{equal}\rput(6.8,3.1){a}
\rput(7,2){$\equal$}\rput(7.2,2.9){b}

\cnode(9,4){0.5}{coveredby}\rput(8.8,4){a}
\pscircle(8.8,4){0.3}\rput(9.3,4){b}
\rput(9,5){$\coveredby$}

\cnode(9,2){0.5}{covers}\rput(8.7,2){a}
\pscircle(9.2,2){0.3}\rput(9.2,2){b}
\rput(9,1){$\covers$}

\cnode(11,4){0.5}{inside}\rput(10.9,4){a}
\pscircle(10.9,4){0.3}\rput(11.35,4){b}
\rput(11,5){$\inside$}

\cnode(11,2){0.5}{contains}\rput(10.65,2){a}
\pscircle(11.1,2){0.3}\rput(11.1,2){b}
\rput(11,1){$\contains$}
\end{pspicture}}
\caption{The eight $\RCC$ relations}
\label{fig:rcc8}
\end{wrapfigure}

\begin{example}[Region Connection Calculus]
\label{ex:rcc8}
The qualitative spatial reasoning with topology introduced
in \cite{randell:1992:computing} and
\cite{egenhofer:1991:reasoning} is concerned with the
following set of qualitative relations:
\[
        \RCC := \{\disjoint,\meet,\overlap,\equal,
                \covers,\contains,\coveredby,\inside\}.
\]
The objects under consideration are here spatial regions,
and each region pair is in precisely one $\RCC$ relation;
see Fig.~\ref{fig:rcc8}.
\end{example}

\vspace{-1ex}
The background knowledge in this case is the set of
possible relation triples pertaining to triples of regions.
For example, the relation triple $\tuple{\meet, \meet,
\meet}$ is possible since there exist three regions
pairwise touching each other. In contrast, the
triple $\tuple{\inside, \inside, \disjoint}$ is impossible
since for any three regions $A,B,C$,
if   $A$ is $\inside$ $B$
and  $B$ is $\inside$ $C$,
then $A$ cannot be $\disjoint$ with $C$.
The set of possible triples is called the
\bfemph{composition table}; it is presented in
the above two papers.
In total, the table lists 193 relation triples.

\begin{example}[Cardinal Directions]
\label{ex:cardinal}
Qualitative reasoning dealing with relative directional
information about point objects
can be formalized using the set of cardinal directions
\[
        \Dir := \{\north, \northeast, \east,
                \southeast, \south, \southwest,
                \west, \northwest,\, \samepoint\},
\]
that consists of the wind rose directions together with the
identity relation denoted by $\samepoint$; see
\cite{frank:1992:qualitative}.  The composition table for
this form of qualitative reasoning is provided in
\cite{ligozat:1998:reasoning}.
\end{example}

\begin{example}[Relative Size]
\label{ex:dir-size}
Qualitative reasoning about relative size of objects is captured
by the relations in the set
\[
        \ms{Size} := \{<, =, >\}.
\]
The corresponding composition table is given in
\cite{gerevini:2002:combining}.
\end{example}

Other examples of qualitative reasoning deal with shape,
directional information about regions
or cyclic ordering of orientations.
In some of them the qualitative
relations are non-binary and the background
knowledge is more complex than the composition table.
To simplify the exposition we assume in the following
binary qualitative relations.


\section{Formalization of Qualitative Reasoning}\label{sec:var-approach}

In what follows we follow the standard terminology of constraint
programming. So by a \bfemph{constraint} on a sequence $\rangeb x1m$
of variables with respective domains $\dom(x_1), \dots, \dom(x_m)$ we
mean a subset of $\dom(x_1) \times \dots \times \dom(x_m)$.  A
\bfemph{constraint satisfaction problem (CSP)} consists of a finite
sequence of variables $X$ with respective domains and a finite set
of constraints, each on a subsequence of $X$\@.  A
\bfemph{solution} to a CSP is an assignment of values to its variables
from their domains that satisfies all constraints.

We study here CSPs with finite domains and solve them using a top-down
search interleaved with constraint propagation.  In our implementation
we use a heuristics-controlled \bfemph{domain partitioning} as the
branching strategy and \bfemph{hyper-arc consistency} of
\cite{mohr:1988:good} as the constraint propagation.

We formalize the qualitative reasoning within the CSP framework as
follows.  We assume a finite set of objects $\OO$, a finite set of binary
qualitative relations $\Q$ and a ternary relation $\CT$ representing
the composition table.  Each qualitative relation between objects
is modelled as a constraint variable the domain of which is a subset
of $\Q$.  We stipulate such a \bfemph{relation variable} for each
ordered pair of objects and organize these variables in an array
$\Rel$ which we call a \bfemph{qualitative array}.

For each triple $a,b,c$ of elements of $\OO$ we have then a
ternary constraint $\comp$ on the corresponding variables:
\begin{multline*}
        \comp(\Rel[a,b],\; \Rel[b,c],\; \Rel[a,c]) :=  \\
        \CT \setintersection
        (\dom(\Rel[a,b]) \times \dom(\Rel[b,c]) \times \dom(\Rel[a,c])).
\end{multline*}

To assume internal integrity of this approach we also adopt
for each ordered pair $a,b$ of elements
of $\OO$, the binary constraint $\conv(\Rel[a,b],\; \Rel[b,a])$
that represents the converse relation table,
and postulate that $\Rel[a,a] = \equal$ for all $a \in \OO$.

We call these constraints \bfemph{integrity constraints}.


\section{Specifying Simulations using Temporal Logic}
\label{sec:constraints}

\begin{wrapfigure}[13]{r}[0mm]{55mm}
\vspace*{-6ex}
\centering
\scalebox{1}{\small\psset{unit=5mm}
\begin{pspicture}(0,-0.5)(10.5,6.5)
\psset{arrows=<->,nodesep=0.2,linewidth=0.5pt}

\pscircle(1,3.7){0.5}
\pscircle(1,2.3){0.5}
\pnode(1.3,3){disjoint}\rput(1,1.2){$\disjoint$}

\pnode(2.6,3){meet1}
\pscircle(3,3.5){0.5}
\pscircle(3,2.5){0.5}
\pnode(3.4,3){meet2}\rput(3,1.5){$\meet$}

\pnode(4.5,3){overlap1}
\pscircle(5,3.3){0.5}
\pscircle(5,2.7){0.5}
\pnode(5.5,3){overlap2}\rput(5,1.7){$\overlap$}

\cnode(7,3){0.5}{equal}
\rput[l](7.6,3){$\equal$}

\cnode(7,5){0.5}{coveredby}
\pscircle(6.8,5){0.3}
\rput(7,6){$\coveredby$}

\cnode(7,1){0.5}{covers}
\pscircle(7.2,1){0.3}
\rput(7,0){$\covers$}

\cnode(9,4){0.5}{inside}
\pscircle(8.9,4){0.3}
\rput(9.2,5){$\inside$}

\cnode(9,2){0.5}{contains}
\pscircle(9.1,2){0.3}
\rput(9.2,1){$\contains$}

\ncline{disjoint}{meet1}
\ncline{meet2}{overlap1}
\ncline{overlap2}{covers}
\ncline{overlap2}{equal}
\ncline{overlap2}{covers}
\ncline{overlap2}{coveredby}
\ncline{equal}{inside}
\ncline{equal}{contains}
\ncline{covers}{contains}
\ncline{coveredby}{inside}
\ncline{equal}{covers}
\ncline{equal}{coveredby}
\end{pspicture}}
\caption{The $\RCC$ neighbourhood relation}
\label{fig:rcc8-neighbours}
\end{wrapfigure}
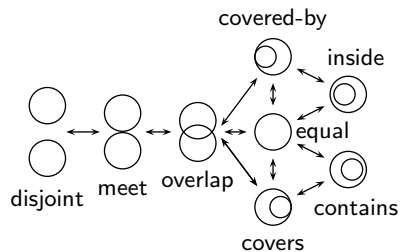

In our framework we assume a \bfemph{conceptual
neighbourhood} between the qualitative relations. This is a
binary relation $\neighbour$
between the elements of the relation set
$\Q$ describing which \emph{atomic} changes in the
qualitative relations are admissible.  So only `smooth'
transitions are allowed. For example, in the case of the
Region Connection Calculus from Example~\ref{ex:rcc8}, the relation
between two regions can change from $\disjoint$ to
$\overlap$ only indirectly via $\meet$. The neighbourhood
relation for $\RCC$ has 22 elements such as
$
\tuple{\disjoint, \meet},
\tuple{\meet, \meet},
\tuple{\meet, \overlap}
$
and their converses and is shown in Fig.~\ref{fig:rcc8-neighbours}.

We assume here that objects can change size
during the simulation.  If we wish to disallow this possibility, then
the pairs $\tuple{\equal, \coveredby}$, $\tuple{\equal, \covers}$,
$\tuple{\equal, \inside}$, $\tuple{\equal, \contains}$ and their
converses should be excluded from the conceptual neighbourhood relation.

In what follows we represent each stage $t$ of
a simulation by a CSP $\PP_t$ uniquely
determined by a qualitative array $Q_t$ and its integrity
constraints.  Here $t$ is a variable ranging over the set
of natural numbers that represents discrete time. Instead
of $Q_t[a,b]$ we also write $Q[a,b,t]$,
as in fact we deal with a \emph{ternary} array.

The stages are linked by \bfemph{inter-state constraints} that
determine which scenarios are allowed. The inter-state constraints
always include constraints stipulating that the atomic changes respect
the conceptual neighbourhood relation. Other inter-state constraints
are problem dependent.

A qualitative simulation corresponds then to a
CSP consisting of \emph{stages} all of which satisfy the integrity
constraints and the problem dependent constraints, and such
that the inter-state constraints are satisfied.
To describe the inter-state constraints we use
\bfemph{atomic formulas} of the form
\[
        Q[a,b] \in \R, \ Q[a,b] \notin \R, \ Q[a,b] = q, \ Q[a,b] \neq q,
\]
where $\R \subseteq \Q$ and $q \in \Q$.
As the latter three forms reduce to the first one, we deal with
the first form only.

We employ a propositional linear temporal logic with four \bfemph{temporal
operators}, $\eventually$ (eventually), $\nexttime$ (next time), $\always$
(from now on) and $\until$ (until), and with the usual connectives.
We use bounded quantification as abbreviations,
\eg, $\phi(o_1) \lor \ldots \lor \phi(o_k)$
abbreviates to $\exists A \in \{\rangeb o1k\}.\,\phi(A)$.

Given a finite set of temporal formulas formalizing the inter-state
constraints we wish then to exhibit a simulation in the form of an
infinite sequence of `atomic' transitions which satisfies these
formulas and respects the integrity constraints.
In the Section~\ref{sec:translations}
we explain how each temporal formula
is translated into a sequence of constraints.

\subsubsection{Paths and loops}

We now proceed by explaining the meaning of a temporal formula
$\phi$ with respect to an arbitrary infinite sequence of qualitative
arrays,
\[
        \pi := Q_1, Q_2, \ldots,
\]
that we call a \bfemph{path}.
Our goal is to implement this semantics, so
we proceed in two stages:
\begin{itemize}
\item First we provide a definition with respect to
        an arbitrary path.
\item Then we limit our attention to specific types of
        paths, which are unfoldings of a loop.
\end{itemize}
In effect, we use here the approach employed in
\emph{bounded model checking}; see \cite{biere:2003:bounded}.
Additionally, to implement this approach in a simple way,
we use a recursive
definition of meaning of the temporal operators instead of
the inductive one.

We write $\models_{\pi} \phi$
to express that $\phi$ holds along the path $\pi$.
We say then that \bfemph{$\pi$ satisfies $\phi$}.
Given $\pi := Q_1, Q_2, \ldots$ we denote by $\pi_i$
the subpath $Q_i, Q_{i+1}, \ldots$. Hence $\pi_1 = \pi$.
The semantics is defined in the standard way, with the
exception that the atomic formulas refer to
qualitative arrays.  The semantics of
connectives is defined independently of the temporal aspect
of the formula.
For other formulas we proceed by recursion as follows:
\[\begin{array}{@{}l@{\hspace{3em}}l@{\hspace{3em}}l}
        \models_{\pi_i} Q[a,b] \in \R &\text{if}&
                Q[a,b,i] \in \R;
                \\[\mnl]
        \models_{\pi_i} \nexttime \phi &\text{if}&
                \models_{\pi_{i+1}} \phi;
                \\[\mnl]
        \models_{\pi} \always \phi &\text{if}&
                \models_{\pi} \phi \text{ and }
                \models_{\pi} \nexttime\always \phi;
                \\[\mnl]
        \models_{\pi} \eventually \phi &\text{if}&
                \models_{\pi} \phi \text{ or }
                \models_{\pi} \nexttime\eventually \phi;
                \\[\mnl]
        \models_{\pi} \chi \until \phi &\text{if}&
                \models_{\pi} \phi \text{ or }
                \models_{\pi} \chi \land \nexttime(\chi \until \phi).
\end{array}\]

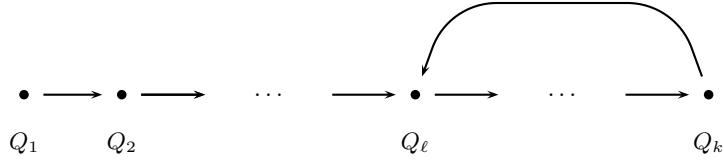
\begin{figure}[t]
\centering
\psset{unit=13mm}
\begin{pspicture}(0.8,0.3)(8.2,2)
\sffamily
\rput(1,0){\dotnode(0,1){t1}\rput(0,0.5){$Q_1$}}
\rput(2,0){\dotnode(0,1){t2}\rput(0,0.5){$Q_2$}}
\rput(3,0){\pnode(0,1){t3}}
\rput(3.5,1){$\ldots$}
\rput(4,0){\pnode(0,1){tlm1}}
\rput(5,0){\dotnode(0,1){tl}\rput(0,0.5){$Q_{\ell}$}}
\rput(6,0){\pnode(0,1){tlp1}}
\rput(6.5,1){$\ldots$}
\rput(7,0){\pnode(0,1){tkm1}}
\rput(8,0){\dotnode(0,1){tk}\rput(0,0.5){$Q_k$}}

\psset{nodesep=0.15}
\ncline{->}{t1}{t2}
\ncline{->}{t2}{t3}
\ncline{->}{t2}{t3}
\ncline{->}{tlm1}{tl}
\ncline{->}{tl}{tlp1}
\ncline{->}{tkm1}{tk}
\ncdiag[angleA=110,angleB=70,arm=0.8,linearc=0.7]{->}{tk}{tl}
\end{pspicture}
\caption{A $(k-\ell)$-loop}
\label{fig:k-l-loop}
\end{figure}

Next, we limit our attention to paths that are loops.
Following \cite{biere:2003:bounded} we call a path
$\pi := Q_1,Q_2,\ldots$ a \bfemph{$(k-\ell)$-loop} if
\[
        \pi = u \cdot v^{*} \qtext{with}
        u := \rangeb Q1{{\ell-1}} \qtext{and}
        v := \rangeb Q{\ell}k;
\]
see Fig.~\ref{fig:k-l-loop}.
By a general result, see \cite{biere:2003:bounded},
for every temporal formula $\phi$ if a path exists that
satisfies it, then a loop path exists that satisfies $\phi$.
This is exploited by our algorithm.
Given a finite set of temporal formulas $\Phi$ it tries to find a path
$\pi := Q_1, Q_2, \ldots$ consisting of qualitative arrays that
satisfies all formulas in $\Phi$, by repeatedly trying to construct an
infinite $(k-\ell)$-loop.  Each such $(k-\ell)$-loop can be finitely
represented using $k$ qualitative arrays.  The algorithm is discussed
in Section~\ref{sec:implementation}.


\section{Temporal Formulas as Constraints}
\label{sec:translations}

A temporal formula restricts the sequence of qualitative arrays
at consecutive stages (time instances).
We now show how to translate these formulas
to constraints in a generic target constraint language.
The translation is based on unravelling the temporal operators
into primitive Boolean constraints and
primitive constraints accessing the qualitative arrays.
Furthermore, we discuss a variation of this translation
that retains more structure of the formula,
using non-Boolean array constraints.

We assume that the target constraint language has primitive
Boolean constraints and reified versions of simple
comparison and arithmetic constraints.
(Recall that a reified constraint generalizes
its base constraint by associating with it a Boolean
variable reflecting its truth.)


\paragraph{Paths with and without loops.}

Both finite and infinite paths can be accommodated within
one constraint model.  To this end, we view a finite sequence of
qualitative arrays together with their integrity
constraints as a single CSP.  The sequence $\rangeb Q1k$
can represent both
\[\begin{array}{@{}ll}
        \text{an infinite path}&
        \pi = (\rangeb Q1{{\ell-1}}) \cdot
                (\rangeb Q{\ell}k)^*, \text{ for some } \ell \geq 1 \text{ and } k \geq \ell,
        \\[\mnl]
        \text{or a finite path}&
        \pi = \rangeb Q1k.
\end{array}\]
To distinguish between these cases, we interpret $\ell$ as a
constraint variable.  We define $\ell = k+1$ to mean that
there is no loop, so we have
$\dom(\ell) = \{1, \ldots, k+1\}$.
A new placeholder array $Q_{k+1}$ is appended
to the sequence of qualitative arrays,
\emph{without} integrity constraints except
the neighbourhood constraints connecting it to $Q_k$.
Finally, possible looping is realized by conditional
equality constraints
\[
        (\ell = j) \ \limplies\  (Q_j = Q_{k+1})
\]
for all $j \in \{1, \ldots, k\}$.
Here
$Q_p = Q_q$ is an equality between qualitative arrays, \ie,
the conjunction of equalities between the corresponding array elements.


\paragraph{Translation into constraints.}

We denote by $\consb i{\phi}b$
the sequence of constraints representing the fact that
formula $\phi$ has the truth value $b$
on the path $\pi_i$.
The translation of a formula $\phi$
on $\rangeb Q1k$
is initiated with $\consb 1{\phi}1$.

We define the constraint translation inductively as follows.
\newcommand{\translationtablespec}%
        {@{}l@{\hspace{1.3em}}l@{\hspace{1.3em}}l}
\newcommand{\nunravel}{n_{\text{unravel}}}
\begin{description}
\item[Atomic formulas:]
        \mbox{}\vspace*{-2mm}
        \[\begin{array}{\translationtablespec}
        \consb i{\ms{true}}b
                &\text{translates to}&
                b = 1;
                \\[\mnl]
        \consb i{Q[a_1,a_2] \in R}b
                &\text{translates to}&
                Q[a_1,a_2,i] = q,
                (q \in R)\equiv b.
        \end{array}\]

\item[Connectives:]
        \mbox{}\vspace*{-3mm}
        \[\begin{array}{\translationtablespec}
        \consb i{\lnot\phi}b
                &\text{translates to}&
                (\lnot b') \equiv b,
                \consb i{\phi}{b'};
        \end{array}\]
        other connectives are translated analogously.
\smallskip

\item[Formula $\nexttime \phi$:]
        The next-time operator takes potential
        loops into account.
        \[\begin{array}{\translationtablespec}
        \consb i{\nexttime \phi}b
                &\text{translates to}
        \\[2\mnl]
        \zerowidth{\hspace{9em}%
        \begin{array}[t]{@{}l}
        \text{if } i < k \text{ then }\\
        \hspace{1em}    \consb {i+1}{\phi}b;\\[\mnl]
        \text{if } i = k \text{ then }\\
        \hspace{1em}%
        \begin{array}{@{}l}
                \begin{array}{@{}l@{}l@{}l}
                \ell = k + 1 &{}\limplies{}& b = 0,\\[\mnl]
                \ell \leq k &{}\limplies{}&
                        b = \lAnd_{j \in \{1, \ldots, k\}}
                        \left(
                        \ell = j \limplies
                        \cons j{\phi}
                        \right).
                \end{array}
        \end{array}
        \end{array}}
        \end{array}\]

\item[Formula $\eventually \phi$:]
        We translate $\eventually \phi$ by unravelling its
        recursive definition
        $\phi \lor \nexttime\eventually \phi$.
        It suffices to do so a finite number
        $\nunravel$ of steps beyond the current
        state, namely the number of steps to reach the loop,
        $\max(0, \ell-i)$, plus the length of the loop,
        $k-\ell$.  A subsequent unravelling step is
        unneeded as it would reach an already visited
        state.  We find
        \begin{align*}
                n_{\text{unravel}} = k - \min(\ell, i).
        \end{align*}
        This equation is a simplification in
        that $\ell$ is assumed constant.  For a variable
        $\ell$, we `pessimistically' replace $\ell$ here
        by the least value
        in its domain, $\mi{min}(\ell)$.
\mbox{}
\item[Formulas $\always \phi$ and $\phi \until \psi$:]
        These formulas are processed analogously to $\eventually \phi$.
\end{description}

The result of translating a formula is a set of primitive reified
Boolean constraints and accesses to the qualitative arrays at certain
times.


\paragraph{Translation using array constraints.}

Unravelling the temporal operators leads to a
creation of several identical copies of subformulas.  In the case of
the $\eventually$ temporal operator where the subformulas
in essence are connected disjunctively, we can do better by
translating differently.
The idea is to push disjunctive information inside the variable domains.
We use \bfemph{array constraints},
which treat array lookups such as
$x = A[\rangen y]$ as a constraint on the variables
$x, \rangen y$ and the (possibly variable) elements of the array $A$.
Array constraints generalize the classic \textsf{element} constraint.

Since we introduce new constraint variables when
translating $\eventually \phi$ using array constraints,
one needs to be careful
when $\eventually\phi$ occurs in the scope
of a negation.  Constraint variables are implicitly
existentially quantified, therefore negation cannot be
implemented by a simple inversion of truth values.
We address this difficulty by first transforming a formula
into a negation normal form, using the standard equivalences
of propositional and temporal logic.

The constraint translations using array constraints (where
different from above) follow.  The crucial difference to
the unravelling translation is that here $i$
is a constraint variable.
\begin{description}
\item[Formula $\eventually \phi$:]
        A fresh variable $j$ ranging over state indices is
        introduced, marking the state at which $\phi$ is
        examined.  The first possible state is the current
        position or the loop start, whichever is earlier.
        Both $\ell$ and $i$ are constraint variables,
        therefore their least possible values
        $\mi{min}(\ell)$, $\mi{min}(i)$, respectively,
        are considered.
        \[\begin{array}{\translationtablespec}
        \consb i{\eventually\phi}b
                &\text{translates to}&
                \text{new } j \text{ with }
                \dom(j) = \{1, \ldots, k\},
                \\[\mnl]&&
                \min( \mi{min}(\ell), \mi{min}(i) ) \leq j,
                \\[\mnl]&&
                \consb j{\phi}b.
        \end{array}\]

\item[Formula $\nexttime \phi$:]
        This case is equivalent to the previous translation of $\nexttime \phi$,
        but we now need to treat $i$ as a variable.
        So  both ``if \ldots{} then'' and $\limplies$ are
        now implemented by Boolean constraints.
\end{description}


\section{Implementation}\label{sec:implementation}

Given a qualitative simulation problem formalized by means of
integrity constraints and inter-state constraints formulated as
temporal formulas, our program generates a solution if one exists or
reports a failure.  During its execution a sequence of CSPs is
repeatedly constructed, starting with a single CSP that is
repeatedly step-wise extended.
The number of steps that need to be considered to conclude
failure depends on the temporal formulas
and is finite \cite{biere:2003:bounded}.
The sequence of CSPs can be viewed as a single finite CSP consisting
of finite domain variables and constraints of a standard type
and thus is each time solvable by generic constraint programming systems.
The top-down search is implemented by means of a regular backtrack
search algorithm based on a variable domain splitting and combined with
constraint propagation.

The variable domain splitting is controlled by
domain-specific heuristics if available.  We make use of
the specialized reasoning techniques due to
\cite{renz:2001:efficient} for $\RCC$ and due to
\cite{ligozat:1998:reasoning} for the cardinal directions.
In these studies maximal tractable subclasses of the
respective calculi are identified and corresponding
polynomial decision procedures for non-temporal qualitative
problems are discussed. In our terminology, if the domain
of each relation variable in a qualitative array belongs to
a certain class, then a certain sequence of domain
splittings intertwined with constraint propagation finds a
solving instantiation of the variables without backtracking
if one exists.  However, here we deal with a more complex set-up:
sequences of qualitative
arrays together with arbitrary temporal constraints
connecting them.  These techniques can then still serve
as heuristics.  We use them in our implementation
to split the variable domains
in such a way that one of the subdomains belongs to a
maximal tractable subclass of the respective calculus.

We implemented the algorithm and both translations of temporal
formulas to constraints in the \eclipse{} constraint programming
system \cite{wallace:1997:eclipse}.
The resulting program is about 2000 lines of code.
We used as constraint propagation hyper-arc consistency
algorithms directly available in \eclipse{} in its \textsf{fd} and
\textsf{propia} libraries and for array constraints
through the implementation discussed in \cite{brand:2001:constraint2}.
In the translations of the temporal formulas,
following the insight from bounded model checking,
redundancy in the resulting generation of constraints
is reduced by sharing subformulas.


\section{Case Study 1: Navigation}

Consider a ship and three buoys forming a triangle. The
problem is to generate a cyclic route of the ship around
the buoys. We reason qualitatively with the cardinal
directions of  Example~\ref{ex:cardinal}.
\begin{itemize}
\item First, we postulate that all objects occupy
        different positions:
        \[
        \always\, \forall a,b \in \OO.\;
        a \neq b \raa Q[a,b] \neq \samepoint.
        \]

\item Without loss of generality we assume that the buoy
        positions are given by
        \[
        \always\, Q[\so{buoy}_a, \so{buoy}_c] = \northwest, \
        \always\, Q[\so{buoy}_a, \so{buoy}_b] = \southwest, \
        \always\, Q[\so{buoy}_b, \so{buoy}_c] = \northwest
        \]
and assume that the initial position of the ship is south of buoy $c$:
        \[
                Q[\so{ship}, \so{buoy}_c] = \south.
        \]

\item To ensure that the ship follows the required
        path around the buoys we stipulate:
        \[
        \always\, \bigl(
        Q[\so{ship}, \so{buoy}_c] = \south \qlimplies
        \begin{array}[t]{@{}l}
        \eventually (Q[\so{ship}, \so{buoy}_a] = \west
        \ \land{} \\[1ex]
        \hspace{1\mindent}\eventually\, (Q[\so{ship}, \so{buoy}_b] = \north
        \ \land{} \\[1ex]
        \hspace{2\mindent}\eventually\, (Q[\so{ship}, \so{buoy}_c] = \east
        \ \land{} \\[1ex]
        \hspace{3\mindent}\eventually\, (Q[\so{ship}, \so{buoy}_c] = \south
        \ ))))
        \bigr).
        \end{array}
        \]
\end{itemize}
In this way we enforce an infinite circling of
the ship around the buoys.

\begin{wrapfigure}[14]{r}[-2mm]{50mm}
\vspace*{-3ex}
\centering
\scalebox{0.87}{\small\psset{unit=5mm}
\begin{pspicture}(0,0)(11,10)
\psset{dimen=outer}
{\psset{linestyle=dotted}
\psline(5,0)(5,10)
\psline(3,0)(3,10)
\psline(8,0)(8,10)
\psline(0,7)(11,7)
\psline(0,5)(11,5)
\psline(0,3)(11,3)
}%

{\psset{fillstyle=solid,linearc=0.1,linewidth=1.6pt}
\pspolygon(2.5,5)(3,4.5)(3.5,5)(3,6)\rput(3,5){$a$}
\pspolygon(4.5,7)(5,6.5)(5.5,7)(5,8)\rput(5,7){$b$}
\pspolygon(7.5,3)(8,2.5)(8.5,3)(8,4)\rput(8,3){$c$}
}%

{\psset{radius=0.5,fillstyle=solid}
\sffamily
\Cnode[linewidth=1.6pt](8,1){p1}\rput(8,1){1}
\Cnode(5,1){p2}\rput(5,1){2}
\Cnode(3,1){p3}\rput(3,1){3}
\Cnode(1,3){p4}\rput(1,3){4}
\Cnode[linewidth=1.6pt](1,5){p5}\rput(1,5){5}
\Cnode(1,7){p6}\rput(1,7){6}
\Cnode(3,9){p7}\rput(3,9){7}
\Cnode[linewidth=1.6pt](5,9){p8}\rput(5,9){8}
\Cnode(8,9){p9}\rput(8,9){9}
\Cnode(10,7){p10}\rput(10,7){10}
\Cnode(10,5){p11}\rput(10,5){11}
\Cnode[linewidth=1.6pt](10,3){p12}\rput(10,3){12}
\Cnode(10,1.5){p13}\rput(10,1.5){13}
}%

{\psset{linestyle=dashed,arrows=->,linewidth=0.5pt,nodesep=3pt}
\ncline{p1}{p2}
\ncline{p2}{p3}
\ncline{p3}{p4}
\ncline{p4}{p5}
\ncline{p5}{p6}
\ncline{p6}{p7}
\ncline{p7}{p8}
\ncline{p8}{p9}
\ncline{p9}{p10}
\ncline{p10}{p11}
\ncline{p11}{p12}
\ncline{p12}{p13}
\ncline{p13}{p1}
}%
\end{pspicture}}
\caption{Navigation path}
\label{fig:navigation}
\end{wrapfigure}
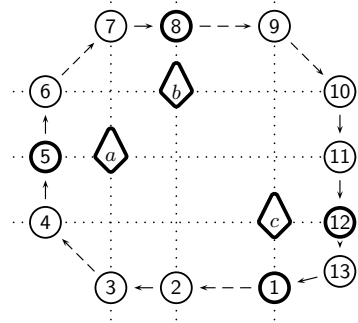

When fed with the above constraints our program generated the infinite
path formed by the cycle through thirteen positions depicted in
Fig.~\ref{fig:navigation}.  The positions required to be visited are
marked by bold circles.  Each of them can be reached from the previous
one through an atomic change in one or more qualitative relations
between the ship and the buoys.
One hour running time was not enough to succeed with the generic
first-fail heuristic, but it took only 20\,s to find the
cycle using the $\Dir$-specific heuristic.
The array constraint translation reduced this slightly to 15\,s.

The cycle found
is a shortest cycle satisfying the specifications. Note that
other, longer cycles exist as well. For example, when starting in
position 1 the ship can first move to an `intermediate' position
between positions 1 and 2, characterized by:
\[
        Q[\so{ship}, \so{buoy}_c] = \southwest, \
        Q[\so{ship}, \so{buoy}_a] = \southeast, \
        Q[\so{ship}, \so{buoy}_b] = \southeast.
\]

We also examined a variant of this problem in which two ships
are required to circle around the buoys while remaining
in the $\north$ or $\northwest$ relation w.r.t.~each other.
In this case the shortest cycle consisted of fifteen positions.


\section{Case Study 2: Simulating of Juggling}

Next, we consider a qualitative formalization of juggling.
We view it as a process
having an initialization phase followed
by a proper juggling phase which is repeated.
As such it fits well our qualitative simulation framework.

We consider two kinds of objects:
the hands and the balls.
For the sake of simplicity,
we only distinguish  the qualitative relations `together',
between a ball and a hand that holds it or between two touching balls,
and `apart'.
This allows us to view the juggling domain
as an instance of an existing topological framework:
we identify `together' and `apart' with
the relations $\meet$ and $\disjoint$ of the $\RCC$ calculus.

In our concrete study,
we assume a single juggler (with two hands) and three balls.
We aim to recreate the three-ball-cascade, see
\cite[p.~8]{gifford:1995:juggling}.
So we have five objects:
\begin{align*}
        \OO &:= \mi{Hands} \setunion \mi{Balls},\\
        \mi{Hands} &:= \{\so{left-hand}, \so{right-hand}\},\\
        \mi{Balls} &:= \setc{\so{ball}_i}{i \in \{1,2,3\}}.
\end{align*}
The constraints are as follows.
\begin{itemize}
\item We only represent the relations of being `together' or `apart':
        \[
        \always\, \forall x,y \in \OO.\;
        (x \neq y \ \limplies\ Q[x,y] \in \{\meet, \disjoint\}).
        \]

\item The hands are always apart:
        \[
        \always\, Q[\so{left-hand}, \so{right-hand}] = \disjoint.
        \]

\item A ball is never in both hands at the same time:
        \[
        \always\, \forall b \in \mi{Balls}.\;
                \neg\left( Q[\so{left-hand}, b] = \meet
                        \ \land\
                        Q[\so{right-hand}, b] = \meet \right).
        \]

\item From some state onwards, at any time instance at most
        one ball is in any hand:
        \[\begin{array}{@{}l}
        \eventually \always\, (
        \forall b \in \mi{Balls}.\;
        \forall h \in \mi{Hands}.\;
        Q[b, h] = \meet
        \ \limplies{}
        \\[\mnl]\hspace{\mindent}
        \forall b_2 \in \mi{Balls}.\;
        b \neq b_2 \ \limplies\
        \forall h_2 \in \mi{Hands}.\;
                Q[b_2, h_2] = \disjoint).
        \end{array}\]

\item Two balls touch if and only if they are in the same hand:
        \[\begin{array}{@{}l}
        \always\, (
                \forall b_1, b_2 \in \mi{Balls}.\;
                b_1 \neq b_2 \ \limplies
                        \\[\mnl]\hspace{\mindent}
                        (Q[b_1, b_2] = \meet \ \lequiv\
                        \exists h \in \mi{Hands}.\;
                                (Q[h, b_1] = \meet \ \land\
                                Q[h, b_2] = \meet))).
        \end{array}\]

\item A ball thrown from one hand remains in the air
        until it lands in the other hand:
        \[\begin{array}{@{}l}
        \always\, (
        \forall b \in \mi{Balls}.\;
        \forall h_1,h_2 \in \mi{Hands}.\;
                h_1 \neq h_2 \land Q[h_1, b] = \meet \limplies
                        \\[\mnl]\hspace{\mindent}
                        \begin{array}{@{}l}
                                Q[h_1, b] = \meet
                                \ \until\
                                \begin{array}[t]{@{}l}
                                        (
                                        Q[h_1, b] = \disjoint \ \land\
                                        Q[h_2, b] = \disjoint \ \land\
                                        \\[\mnl]\
                                        (Q[h_1, b] = \disjoint
                                        \ \until\
                                        Q[h_2, b] = \meet)
                                        )).
                                \end{array}
                        \end{array}
        \end{array}\]

\item A ball in the air will land before any
        other ball that is currently in a hand,
        \[\begin{array}{@{}l}
        \always\, (
        \forall h_1, h_2 \in \mi{Hands}.\;
        \forall b_1, b_2 \in \mi{Balls}.\;
        Q[h_1, b_1] = \disjoint \land
        Q[h_2, b_2] = \meet \limplies{}
        \\[\mnl]\hspace{2\mindent}
        Q[h_2, b_2] = \meet \until
        ((\forall h \in \mi{Hands}.\; Q[h, b_2] = \disjoint)
        \\[\mnl]\hspace{7\mindent}
        {} \until (\exists h \in \mi{Hands}.\; Q[h, b_1] = \meet))).
        \end{array}\]

\item No two balls are thrown at the same time:
        \[\begin{array}{@{}l}
        \always\, (\forall b_1,b_2 \in \mi{Balls}.\;
        b_1 \neq b_2 \limplies
        \forall h_1,h_2 \in \mi{Hands}.\\[\mnl]
        \hspace{\mindent}\neg(
                \begin{array}[t]{@{}l}
                        Q[b_1, h_1] = \meet \land
                        \nexttime\, Q[b_1, h_1] = \disjoint
                \ \land{}\\[\mnl]
                        Q[b_2, h_2] = \meet \land
                        \nexttime\, Q[b_2, h_2] = \disjoint )).
                \end{array}
        \end{array}\]

\item A hand can interact with only one ball at a time:
        \[\begin{array}{@{}l}
        \always\,
        \forall h  \in \mi{Hands}.\;
        \forall b_1 \in \mi{Balls}.\;
        \\[\mnl]\hspace{\mindent}
                (
                \begin{array}[t]{@{}l}
                Q[h, b_1] = \meet \ \land\
                \nexttime\, Q[h, b_1] = \disjoint
                \ \lor{}\\[\mnl]
                Q[h, b_1] = \disjoint \ \land\
                \nexttime\, Q[h, b_1] = \meet
                )
                \ \limplies {}\\[\mnl]
                \end{array}
                \\[\mnl]\hspace{2\mindent}
                \forall b_2 \in \mi{Balls}.\;
                b_1 \neq b_2 \ \limplies
                \\[\mnl]\hspace{3\mindent}
                        \begin{array}[t]{@{}l}
                        (Q[h, b_2] = \meet \ \limplies\
                        \nexttime\, Q[h, b_2] = \meet)
                        \ \land{}\\[\mnl]
                        (Q[h, b_2] = \disjoint \ \limplies\
                        \nexttime\, Q[h, b_2] = \disjoint).
                        \end{array}
        \end{array}\]
\end{itemize}

\noindent
Initially balls 1 and 2 are in the left hand,
while ball 3 is in the right hand:
\[
        Q[\so{left-hand},  \so{ball}_1] \!=\! \meet,
        Q[\so{left-hand},  \so{ball}_2] \!=\! \meet,
        Q[\so{right-hand}, \so{ball}_3] \!=\! \meet.
\]

\begin{figure}[b]
\hspace*{-5mm}
\scalebox{0.885}{%
\footnotesize
\psset{unit=5.7mm}
\begin{pspicture}(0,-1)(25,6)

\newcommand{\ball}[2]{\pscircle(#2){0.27}\rput(#2){#1}}

\newcommand{\balllu}[2]{\ball{#1}{#2}\rput(#2)
	{\pscurve[linestyle=dotted]{->}(-1.1,-1)(-0.8,-0.3)(-0.4,-0.1)}}

\newcommand{\ballru}[2]{\ball{#1}{#2}\rput(#2)
	{\pscurve[linestyle=dotted]{->}(1.1,-1)(0.8,-0.3)(0.4,-0.1)}}

\newcommand{\ballld}[2]{\ball{#1}{#2}\rput(#2)
	{\pscurve[linestyle=dotted]{<-}(-1.1,-1)(-0.8,-0.3)(-0.4,-0.1)}}

\newcommand{\ballrd}[2]{\ball{#1}{#2}\rput(#2)
	{\pscurve[linestyle=dotted]{<-}(1.1,-1)(0.8,-0.3)(0.4,-0.1)}}

{\psset{linewidth=0.3pt}\multips(0,0)(3,0){9}{\qline(0,1.1)(0,5.5)}}%
\psline[linewidth=0.5pt,doubleline=true](6,1.0)(6,5.6)

{\rmfamily\newcounter{juggaxiscounter}
\multiput(0,0)(3,0){8}{\addtocounter{juggaxiscounter}{1}
\rput(1.5,1){State \arabic{juggaxiscounter}}}}%

{\psset{linewidth=0.5pt}\sffamily\footnotesize

\rput(0,2){\pnode(1.5,-1.3){state1}
\ball1{0.33,0}
\ball2{0.87,0}
\ball3{2.5,0}
}%

\rput(3,2){\pnode(1.5,-1.3){state2}
\balllu1{1.5,3}
\ball2{0.5,0}
\ball3{2.5,0}
}%

\rput(6,2){\pnode(1.5,-1.3){state3}
\ballru3{1.4,2.5}
\ballrd1{1.6,3.2}
\ball2{0.5,0}
}%

\rput(9,2){\pnode(1.5,-1.3){state4}
\balllu2{1.4,3.2}
\ballld3{1.6,2.5}
\ball1{2.5,0}
}

\rput(12,2){\pnode(1.5,-1.3){state5}
\ballrd2{1.6,3.2}
\ballru1{1.4,2.5}
\ball3{0.5,0}
}

\rput(15,2){\pnode(1.5,-1.3){state6}
\ballld1{1.6,2.5}
\balllu3{1.4,3.2}
\ball2{2.5,0}
}%

\rput(18,2){\pnode(1.5,-1.3){state7}
\ballrd3{1.6,3.2}
\ballru2{1.4,2.5}
\ball1{0.5,0}
}%

\rput(21,2){\pnode(1.5,-1.3){state8}
\ballld2{1.6,2.5}
\balllu1{1.4,3.2}
\ball3{2.5,0}
}}%

{\psset{angleA=335,angleB=205,nodesep=0.6em,linewidth=1.2pt}
\nccurve{->}{state1}{state2}
\nccurve{->}{state2}{state3}
\nccurve{->}{state3}{state4}
\nccurve{->}{state4}{state5}
\nccurve{->}{state5}{state6}
\nccurve{->}{state6}{state7}
\nccurve{->}{state7}{state8}
\ncdiag[angleA=290,angleB=250,arm=1,linearc=1]%
{<-}{state3}{state8}
}%

\end{pspicture}}
\caption{Simulation of Juggling}
\label{fig:juggling}
\end{figure}
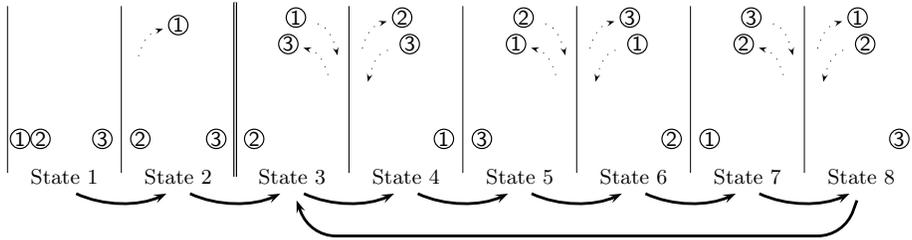

Note that the constraints enforce that the juggling
continues forever. Our program finds an infinite
simulation in the form of a path $[1..2][3..8]^*$;
see Fig.~\ref{fig:juggling}.
The running time was roughly 100\,s using the
generic first-fail heuristic;
the $\RCC$-specific heuristic, resulting in 43\,min,
was not useful.

We stress the fact that the complete specification of this problem is
not straightforward. In fact, the interaction with our program
revealed that the initial specification was incomplete.  This led us to
the introduction of the last constraint.


\subsection*{Aspect Integration: Adding Cardinal Directions}

The compositional nature of the `relations as variables' approach
makes it easy to integrate several spatial aspects (\eg, topology
\emph{and} size, direction, shape etc.) in one model.  For the
non-temporal case, we argued in \cite{brand:2004:relation} that the
background knowledge on linking different aspects can be viewed as
just another integrity constraint.  Here we show that also qualitative
simulation and aspect integration combine easily, by extending the
juggling example with the cardinal directions.

As the subject of this paper is modelling and solving,
not the actual inter-aspect background knowledge,
we only explain the integration of the three relations
$\meet, \disjoint, \equal$ with the cardinal directions $\Dir$.
We simply add
\[
        \link(Q[a,b],\, Q_{\Dir}[a,b]) \ := \
                (Q[a,b] = \equal) \lequiv
                (Q_{\Dir}[a,b] = \samepoint)
\]
as the aspect linking constraint.
It refers to the two respective qualitative arrays and
is stated for all spatial objects $a,b$.
We add the following domain-specific requirements
to our specification of juggling:
        \[\begin{array}{@{}l}
        Q_{\Dir}[\so{left-hand}, \so{right-hand}] = \west;
        \\[1.5\mnl]
        \always\,
        \forall b \in \mi{Balls}.\;
        \forall h \in \mi{Hands}.\;
                Q[b,h] = \meet \limplies Q_{\Dir}[b,h] = \north;
        \\[1.5\mnl]
        \always\,\forall b \in \mi{Balls}.\;
                Q[b,\so{left-hand}] = \disjoint \land
                Q[b,\so{right-hand}] = \disjoint
                \limplies\\[\mnl]\hspace{4\mindent}
                Q_{\Dir}[b,\so{left-hand}] \neq \north \land
                Q_{\Dir}[b,\so{right-hand}] \neq \north.
        \end{array}\]
We state thus that a ball in a hand is `above' that hand,
and that a ball is not thrown straight upwards.

This simple augmentation of the juggling domain with
directions yields the same first simulation
as in the single-aspect case,
but now with the $\RCC$ and $\Dir$ components.
The ball/hand relation just alternates between $\north$ and $\northwest$
(or $\northeast$).

We emphasize that it was straightforward to extend
our implementation to achieve the integration of two aspects.
The constraint propagation for the $\link$ constraints
is achieved by the same generic hyper-arc consistency algorithm
used for the single-aspect integrity constraints.
This is in contrast to the `relations as constraints' approach
which requires new aspect integration \emph{algorithms};
see, \eg, the bipath-consistency algorithm of \cite{gerevini:2002:combining}.


\section{Conclusions}
\label{sec:conclusions}

\subsubsection*{Related Work}

The most common approach to qualitative simulation is
the one discussed in \cite[Chapter 5]{kuipers:1994:qualitative}.
For a recent overview see \cite{kuipers:2001:encyclopedia}.
It is based on a qualitative differential equation model
(QDE) in which one abstracts from the usual differential
equations by reasoning about a finite set of symbolic values
(called \emph{landmark values}).
The resulting algorithm, called \emph{QSIM}, constructs
the tree of possible evolutions by repeatedly constructing the
successor states. During this process CSPs are generated and solved.
This approach is best suited to simulate the evolution of
physical systems.

Our approach is inspired by the qualitative spatial simulation studied
in \cite{cui:1992:qualitative}, the main features of which are
captured by the composition table and the neighbourhood relation
discussed in Example~\ref{ex:rcc8}.  The distinction between the
integrity and inter-state constraints is introduced there; however, the
latter only link consecutive states in the simulation. As a
result, our case studies are beyond their reach.
Our experience with our program moreover suggests that
the algorithm of \cite{cui:1992:qualitative} may not be
a realistic basis for an efficient qualitative reasoning system.

To our knowledge the `(qualitative) relations as variables' approach
to modelling qualitative reasoning was first used in
\cite{tsang:1987:consistent}, to deal with the qualitative temporal
reasoning due to \cite{allen:1983:maintaining}.  In
\cite{renz:2001:efficient} this approach is used in an argument to
establish the quality of a generator of random scenarios, whilst the
main part of this paper uses the customary `relations as constraints'
approach.  In \cite[pages 30-33]{apt:2003:principles} we applied the
`relations as variables' approach to model a qualitative spatial
reasoning problem.  In \cite{brand:2004:relation} we used it to deal
in a simple way with aspect integration and in \cite{AB05} to study
qualitative planning problems.

In \cite{Brz91} various semantics for a programming language
that combines temporal logic operators with constraint logic programming are studied.
Finally, in the
\textsc{TLPlan} system of \cite{bacchus:2000:using} temporal logic
is used to support the construction of control rules that guide plan
search.  The planning system is based on an incremental forward-search,
so the temporal formulas are unfolded one step at a time, in contrast
to the translation into constraints in our constraint-based system.

\subsubsection{Summary}

We introduced a constraint-based framework for describing infinite
qualitative simulations.  Simulations are formalized by means of
inter-state constraints that are defined using linear temporal logic.
This results in a high degree of expressiveness.  These constraints are
translated into a generic target constraint language.  The qualitative
relations are represented as domains of constraint variables.  This
makes the considered CSPs finite, allows one to use hyper-arc
consistency as constraint propagation, and to integrate various
knowledge aspects in a straightforward way by simply adding linking
constraints.

We implemented this approach in a generic constraint programming system,
\eclipse, using techniques from bounded model checking and
by experimenting with various heuristics.
The resulting system is conceptually simple and easy to use and
allows for a straightforward modification of the problem specifications.
We substantiated these claims by means of two detailed case studies.


\bibliographystyle{plain}

\end{document}